\documentclass[runningheads]{llncs}
\usepackage[T1]{fontenc}
\usepackage{graphicx}
\usepackage{amsmath}
\usepackage{amssymb} 
\usepackage{subcaption} 
\usepackage{multirow}
\usepackage{soul, color, xcolor}

\begin{document}

%
%
\title{Brain-inspired Action Generation with Spiking Transformer Diffusion Policy Model}

\author{Qianhao Wang\inst{1,2} \and
Yinqian Sun\inst{1} \and
Enmeng Lu\inst{1} \and
Qian Zhang\inst{1,2,4} \thanks{Corresponding author. Email: q.zhang@ia.ac.cn } \and
Yi Zeng\inst{1,2,3,4} \thanks{Corresponding author. Email: yi.zeng@ia.ac.cn}
}

\institute{Brain-inspired Cognitive Intelligence Lab, Institute of Automation, Chinese Academy of Sciences, Beijing, China.  \and
School of Artiﬁcial Intelligence, University of Chinese Academy of Sciences,Beijing, China.  \and
Key Laboratory of Brain Cognition and Brain-inspired Intelligence Technology, CAS, Shanghai, China. \and 
Center for Long-term Artiﬁcial Intelligence, Beijing, China.
\\}
\maketitle              
%

\begin{abstract}
Spiking Neural Networks (SNNs) has the ability to extract spatio-temporal features due to their spiking sequence. While previous research has primarily foucus on the classification of image and reinforcement learning. In our paper, we put forward novel diffusion policy model based on Spiking Transformer Neural Networks and Denoising Diffusion Probabilistic Model (DDPM): Spiking Transformer Modulate Diffusion Policy Model (STMDP), a new brain-inspired model for generating robot action trajectories. In order to improve the performance of this model, we develop a novel decoder module: Spiking Modulate Decoder (SMD), which replaces the traditional Decoder module within the Transformer architecture. Additionally, we explored the substitution of DDPM with Denoising Diffusion Implicit Models (DDIM) in our framework. We conducted experiments across four robotic manipulation tasks and performed ablation studies on the modulate block. Our model consistently outperforms existing Transformer-based diffusion policy method. Especially in Can task, we achieved an improvement of 8$\%$. The proposed STMDP method integrates SNNs, dffusion model and Transformer architecture, which offers new perspectives and promising directions for exploration in brain-inspired robotics.

\end{abstract}
\section{Introduction}
Spiking Neural Networks transmit information through spike sequences, which encode both temporal and spatial information. This grants SNNs spatio-temporal feature extraction capabilities \cite{qiao2021survey}. The most common models include the Integrate-and-Fire (IF) neuron, the Leaky Integrate-and-Fire (LIF) neuron \cite{gerstner2002spiking}, and the Hodgkin-Huxley (HH) neuron \cite{hodgkin1952quantitative}. Since SNNs are modeled in a more biologically plausible manner, their training processes differ slightly from those of artificial neural networks (ANNs).

Behavior Cloning (BC), a branch of imitation learning, directly learns a policy by mapping states to actions based on expert demonstration trajectories \cite{wu2019imitation}. But, it suffers from compounding errors, resulting in poor generalization and suboptimal performance in practical scenarios. Action Chuncking with Transformers (ACT) \cite{zhao2023learning} employs a Variational Autoencoder (VAE) for imitation learning, generating action trajectories, while diffusion policy \cite{chi2023diffusion} uses diffusion models for the same purpose. Due to the current transformer-based spiking neural network models performing poorly in generating action trajectories. We propose the Spiking Transformer Modulate Diffusion Policy Model and the Spiking Modulate Decoder to enhance the accuracy of robotic manipulation tasks.

In this work, we build upon the diffusion policy method \cite{chi2023diffusion} and integrate it with SNNs to tackle the aforementioned challenges. Diffusion policies offer strong generative capabilities, enabling them to learn diverse patterns from data, while SNNs’ spatio-temporal feature extraction ensures more coherent action trajectories. We propose a novel diffusion policy model based on Spiking Transformer Neural Networks, along with a novel decoder module.

Our key contributions are as follows:
\begin{itemize}
\item[$\bullet$] We present the Spiking Transformer Modulate Diffusion Policy Model , and its core module of this model: the Spiking Modulate Decoder module.
\item[$\bullet$] Additionally, We implemented diffusion policies utilizing a spiking Transformer architecture, and develop the Spiking Diffusion Transformer (SDIT), which is based on Diffusion with Transformer (DIT) framework.
\item[$\bullet$] We conducted experiments on four robotic manipulation tasks, demonstrating that our proposed STMDP model significantly enhanced performance. Specifically, in the Can task, our model outperformed the current best Transformer-based Diffusion Policy model by 8\%. 
\end{itemize}

\section{Related Work}

SNNs are composed of various spiking neuron models, each differing in their complexity of neuron modeling. SNNs have been widely applied in image processing \cite{wang2023masked,zhou2022spikformer} and reinforcement learning \cite{zhang2022multi,chen2024fully,oikonomou2023hybrid}. Common training methods include Spike-Timing-Dependent Plasticity (STDP) \cite{bi1998synaptic,song2000competitive}, ANN-to-SNN conversion \cite{rueckauer2017conversion,sengupta2019going}, and direct training methods \cite{lee2020enabling,neftci2019surrogate}. In our study, we utilized the LIF neuron model and the direct training method. Cao \cite{cao2024spiking} introduced the SDDPM, which improved image generation with low energy consumption. Hou \cite{hou2024sdp} proposed the SDP, enhancing robotic manipulation performance. While their work is based on the Unet architecture, our approach uses a transformer architecture. Furthermore, we introduce a more general module: SMD.

BC is attractive due to its simplicity, easy of training, and interpretability. However, it suffers from compounding errors (distributional drift and cumulative errors) and limited behavior patterns \cite{bansal2018chauffeurnet}. To address these limitations, numerous methods have been proposed, such as data augmentation to enhance data diversity and mitigate compounding errors \cite{bojarski2016end}, and the use of mixture Gaussian models \cite{mandlekar2021learning,shafiullah2022behavior} or implicit models \cite{wu2020spatial} to capture more diverse behavior patterns.

The Denoising Diffusion Probabilistic Model (DDPM) \cite{ho2020denoising} improves generative performance by predicting noise instead of the noisy image itself, reducing the complexity of prediction. DDIM \cite{song2020denoising} constructs a posterior distribution such that its marginal distribution is Gaussian, eliminating the need for the forward process to follow a Markov chain. Recent work has explored replacing the typical Unet-based diffusion model with transformer-based diffusion models, yielding better generative results \cite{peebles2023scalable}. Our work differs in that we modify the decoding module rather than the encoding module, allowing for more flexible handling of conditional inputs, and our model is primarily designed for action trajectory generation instead of image generation. 

\section{Method}

\subsection{Preliminaries}
\textbf{Spiking Neural Networks:} SNNs imitate the signal transmission process in the brain and transmit information in sparse spike sequences. We use the LIF neuron model for modeling, which is suitable for deep learning tasks. The dynamic process can be described as Equation \eqref{eq1}. In practical applications, we discretize it to facilitate forward process and training. It is described as Equation \eqref{eq2} and Equation \eqref{eq3}.
\begin{equation}
\label{eq1}
	\tau\frac{\mathrm{d} V(t)}{\mathrm{d} t} = -(V(t)-V_{reset})+I(t)
\end{equation}

\begin{equation}
\label{eq2}
	 u_{t+1,n+1}^i =ku_{t,n+1}^i(1-o_{t,n+1}^i)+\sum_{j=1}^{l(n)} w_n^{i j}o_{t,n+1}^j
\end{equation}

\begin{equation}
\label{eq3}
	 o_{t+1,n+1}^i = f(u_{t+1,n+1}^i-V_{th})
\end{equation}
where $I(t)$ represents the incoming synaptic current, $V(t)$ denotes the membrane potential, $\tau$ is the time constant, $V_{reset}$ is the resetting potential. When the membrane potential exceeds the threshold $V_{th}$, the neuron fires a spike, and the membrane potential is reset to the resetting potential.

The training method of our model uses direct training, the derivation process of direct training is described in Equation\eqref{eq4} and Equation \eqref{eq5}.

\begin{equation}
\label{eq4}
	\frac{\partial L}{\partial o_{t,n}^i} =\sum_{j=1}^{l(n+1)} \frac{\partial L}{\partial o_{t,n+1}^i} \frac{\partial o_{t,n+1}^i}{\partial o_{t,n}^i}+\frac{\partial L}{\partial o_{t+1,n}^i} \frac{\partial o_{t+1,n}^i}{\partial o_{t,n}^i}
\end{equation}

\begin{equation}
\label{eq5}
	\frac{\partial L}{\partial u_{t,n}^i} = \frac{\partial L}{\partial o_{t,n}^i} \frac{\partial o_{t,n}^i}{\partial u_{t,n}^i}+\frac{\partial L}{\partial o_{t+1,n}^i} \frac{\partial o_{t+1,n}^i}{\partial o_{t,n}^i}
\end{equation}
where $n$ represents the number of layers in the network, $t$ denotes the timestep, $i$ refers to the $neuron_i$ in a given layer, $k$ is a constant. $f(x)$ is a step function, $V_{th}$ represents the threshold value, $L$ denotes the loss function, and $l(n)$ indicates the number of neurons. $o_{t,n}^i$ represents the output(0/1) of the $i_{th}$ neuron in the $n_{th}$ layer at the $t_{th}$ timestep. $u_{t,n}^i$ represents the membrane potential of the neuron and $w_n^{i j}$ denotes the weight.

Becaues of the $f(x)$ is undifferentiable ,we use a surrogate function $g(x)$ to solve this problem during the back-propagation process. The surrogate function is described as Equation\eqref{eq6}.

\begin{equation}
\label{eq6}
	g(x) = 
        \begin{cases}
        0, & x\leq-\frac{1}{\epsilon},\\
        -\frac{1}{2}\epsilon^2\vert x\vert, & x \textless \vert \frac{1}{\epsilon} \vert ,\\
        1, & x\geq \frac{1}{\epsilon}\\
        \end{cases}
\end{equation}
where $g(x)$ represents the surrogate function, $\epsilon$ is a constant, we set $\epsilon = 2$.

\textbf{DDPM and DDIM:} We constructed STMDP spiking neural based DDPM and also explored replacing DDPM with DDIM. Before introducing the architectures, we briefly review some diffusion model concepts involved in the research. 
The diffusion model includes a forward process $q(x_t \mid x_0) = \mathcal{N}(x_t;\sqrt{\overline{\alpha_t}}x_0,(1-\overline\alpha_t)I)$, which adds noise to the data; and a reverse process $q_\theta(x_{t-1}\mid x_t) = \mathcal{N}(\mu_\theta(x_t),\Sigma_\theta(x_t))$, which recovers data from random noise. Diffusion models are trained to learn $q_\theta$, using the variational lower bound\cite{kingma2013auto} for training, simplified to minimize the KL divergence between $q_\theta$ and $q(x_{t-1}\mid x_t,x_0)$. This paper employs a fixed variance $\epsilon_\theta(x_t)=\beta I$. Finally, the predicted mean is converted to predicted noise through the reparameterization process $L_{t-1}(\theta)= \|\epsilon_{\theta}(x_t)-\epsilon\|^2$.The DDPM forward process is a Markov process, whereas DDIM eliminats the dependency on the forward Markov process and can generate high-quality samples in fewer sampling timesteps.

\begin{figure*}[t!]
	\includegraphics[width=\textwidth]{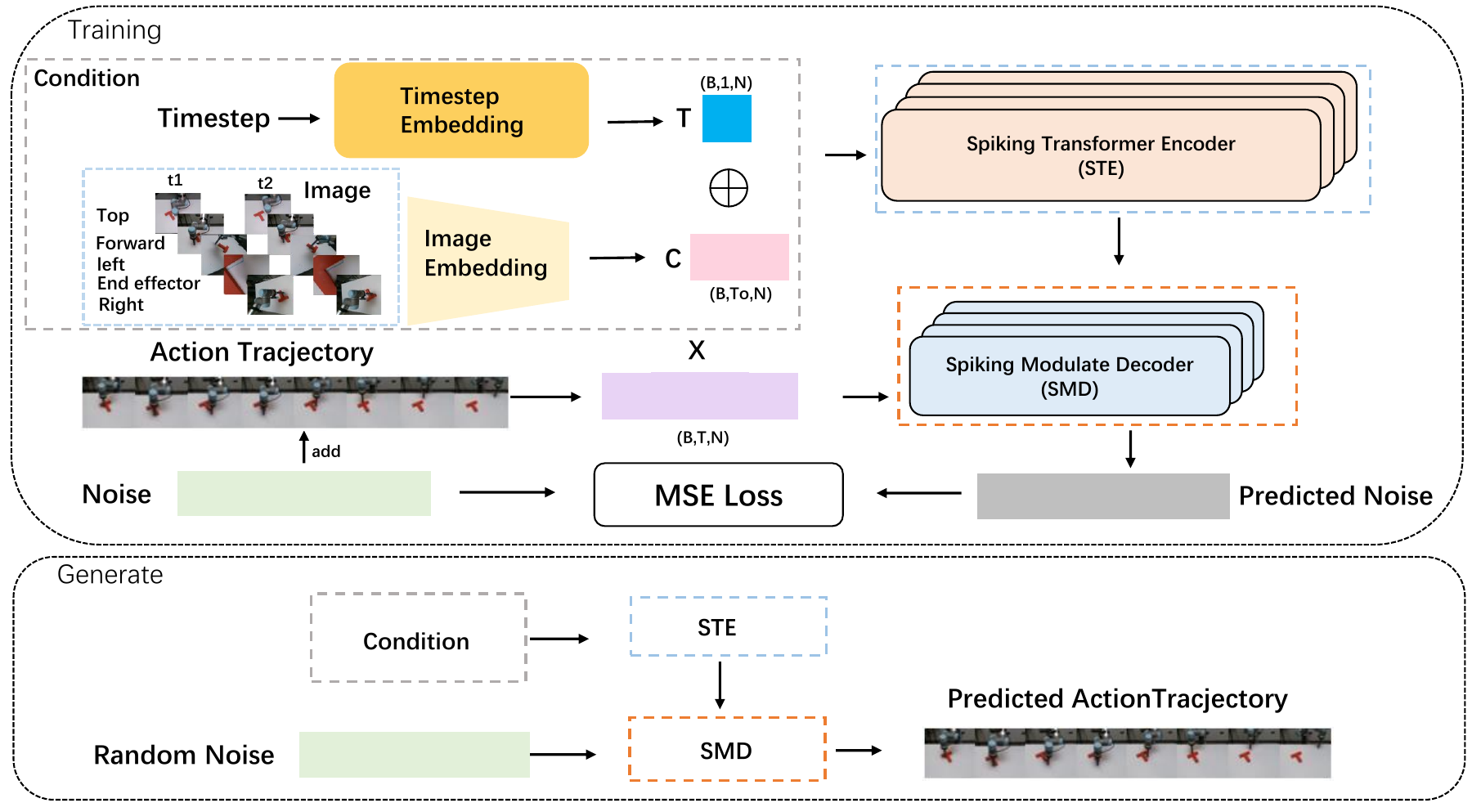} %
	\caption{The architect of spiking transformer modulate diffusion policy module. }
	\label{fig:STDPM_arc}
\end{figure*}

\subsection{Spiking Transformer Modulate Diffusion Policy} 

The STMDP model differs from the DIT architecture, which is primarily based on the Transformer encoder structure \cite{vaswani2017attention}. In contrast, our proposed model preserves the encoder-decoder architecture of the Transformer. Additionally, STMDP is built upon a spiking neural network.
The encoding strategy involves direct encoding, where the input is replicated across eight time steps, and decoding is performed by averaging the inputs over these steps to generate the output. The overall architecture of the model is illustrated in Figure \ref{fig:STDPM_arc}.


The diffusion policy process consists of two phases: a training phase and a generation phase. During the training phase, a noise prediction model is trained, and in the generation phase, the trained model is employed to progressively reduce noise and generate robot-executable action trajectories guided by visual input. For the training phase, a fixed time step is embedded to high dimension vector as $T$. Image frames are extracted from demonstration videos, and multi-angle images are encoded as high dimension vector $C$. The timestep embedding and image embedding are then combined and input into the Spiking Transformer Encoder (STE) module, followed by processing in the SMD module for guidance. After adding noise to the action trajectory, it is encoded into the X vector and decoded by the SMD module  together with the output of STE. The decoder module outputs a prediction of the noise in the action trajectories, which is updated using the Mean Square Error (MSE) with target noise $Y$. During the generation phase, images and inference time steps serve as conditions; however, the input to the SMD module is random noise, which is iteratively denoised to produce robot-executable action trajectories.
This can be represented:

\begin{equation}
\label{eq7}
	 M = SSA(T,C)+FFN(SSA(T,C))
\end{equation}
\begin{equation}
\label{eq8}
	L= \|Y-SMDM(M,X)\|^2
\end{equation}
where $SSA$ stands for Spiking Self Attention module and $FFN$ stands for Feed Forward neural Network module and $M$ stands for the output of the STE module. $Y$ is noise and $SMDM$ stands for SMD module.

\subsection{Base Block}

\begin{figure}[htbp]
    \centering
	\includegraphics[width=0.7\textwidth]{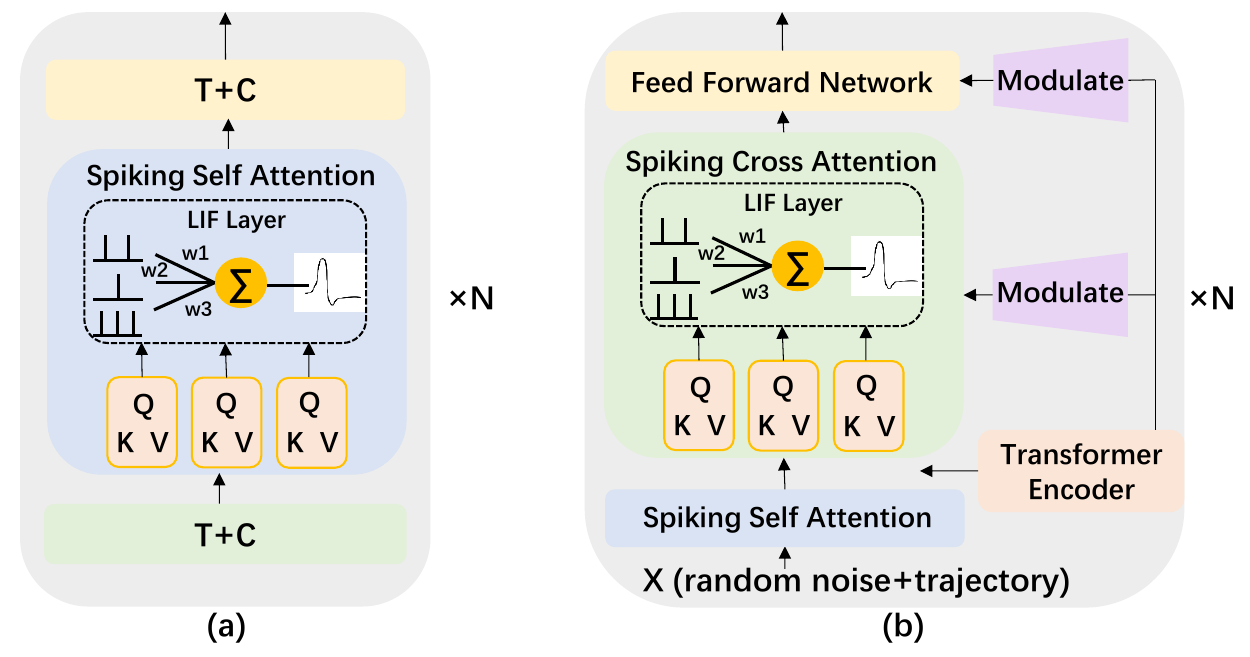} %
	\caption{The architect of base block.(a) represents the architecture of Spiking Encoder;(b) represents the architecture of SMD. }
	\label{fig:encoder_decoder}
\end{figure}

In the Spiking Encoder Module, the core component is the spiking multi-head self-attention block, which combines a LIF layer with a multi-head self-attention mechanism. The remaining blocks retain the fundamental structure of the standard encoder block. In the SMD module, the base block includes a spiking multi-head self-attention block, a spiking modulated cross-attention block, and a modulated feed-forward network block. The detailed structures of the basic blocks within the Spiking Encoder and SMD modules are depicted in Figures \ref{fig:encoder_decoder}, respectively. A key difference in the SMD module is that modulation is applied to the cross-attention module, contrasting with the DIT structure, which primarily modulates the self-attention mechanism in the encoder. This design is inspired by the architecture of BERT, which is based on encoder modules, and GPT, which is built on decoder modules—both of which have demonstrated strong generative capabilities. 

\section{Experiment}
\subsection{Datasets and Experimental Settings}

\begin{figure}[htbp]
	\centering
	\includegraphics[width=0.7\textwidth]{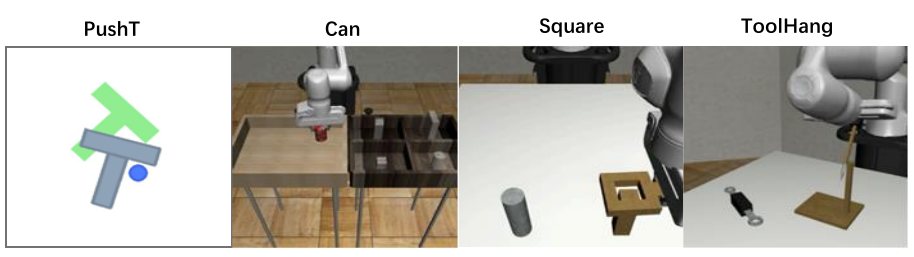} %
	\caption{Tasks used for the experiment. }
	\label{fig:experiment}
\end{figure}

The proposed STMDP model is evaluated across four robotic manipulation tasks: PushT, Can, Square and ToolHang (Square and Can is mh version, ToolHang is ph verseion). These tasks cover a diverse range of actions, including pushing and grasping, as shown in Figure \ref{fig:experiment}, offering a thorough assessment of the model’s performance. PushT: Move the T-shaped block to the specified position on the desktop; Can: Grasping a can of soda and placing it into a slot; Square: Lifting an object with a square hole and fitting it onto a square pillar; ToolHang: Assembling a rack from tools on the table and hanging the remaining tools on the rack.

We present results derived from the average of the last 10 checkpoints (saved every 50 epochs) across 50 initializations of the environment, totaling an average of 500 experiments, while all other settings remain consistent with the previous configurations\cite{chi2023diffusion}. When conducting experiments with DDIM model, we adopt the experimental setups: the value of $\alpha$ is set to 0 to make the forward process of its diffusion process deterministic, the inference timestep is set to 50 and is randomly selected (in contrast to 100 in DDPM).

\subsection{Experiment Results}
To ensure a comprehensive comparison, the experiments have implemented diffusion policies based on spiking Transformer (SDPT), which utilizing the same Transformer architecture as the diffusion policy based on ANN described (DPT) in \cite{chi2023diffusion}, and those based on spiking DIT (SDIT) structures as in~\cite{peebles2023scalable}. The main difference between them is whether there is Spiking Transformer (ST).
The Modulate Block (M-Block) plays a critical role in the DIT model. This work also implements a spiking modulate decoder module based on the M-Block and experimentally validates its effectiveness. Additionally, we replace DDPM in our model with DDIM and propose STMDP-I to conduct experiments. The experimental results are shown in Table \ref{table1}. The STMDP achieved the 75.4\% in PushT task, it exceeded the previous best result by nearly 5\% (achieved by DPT). As for the STMDP-I, We guess that because the parameters $\alpha$ and inference step are not selected appropriately. While It achieved the best result in the Square task, demonstrating its potential.

\begin{table*}[htbp]
\centering
\caption{Execution results of all models on four tasks. }
\label{table1}
\begin{tabular}{c|c|c|c|c|c|c}
\hline
\multirow{2}{*}{Model} &\multicolumn{2}{c|}{Blocks} & \multicolumn{4}{c}{Tasks}\\
\cline{2-7}

 &ST &M-Block & PushT & Can & Square & ToolHang \\
\hline

DPT\cite{chi2023diffusion} &&&	0.707 & 0.860	&0.720	&0.540\\ 
DIT\cite{peebles2023scalable} &&\checkmark&	0.657 & 0.920	&0.800	&\textbf{0.780}\\ 
\hline
SDPT     &\checkmark&&	0.747 & 0.220	&0.040	&0.000\\ 
SDIT   &\checkmark&\checkmark&	0.700 & 0.920	&0.660	&0.540\\
\hline
\textbf{STMDP}  &\checkmark&\checkmark& \textbf{0.754} & \textbf{0.940}	& 0.800	&0.540\\ 
\textbf{STMDP-I}  &\checkmark&\checkmark& 0.640 & 0.740	& \textbf{0.880}	&0.720\\ 
\hline
\end{tabular}
\label{table_1}
\end{table*}

\begin{figure}[htbp]
	\centering
	\includegraphics[width=0.4\textwidth]{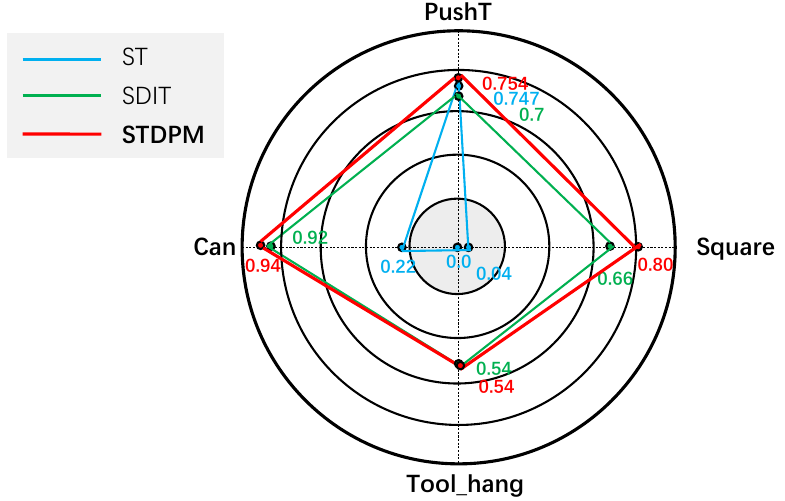} %
	\caption{The results of the ablation study. }
	\label{fig4}
\end{figure}

\textbf{Impact of the Modulate Blocks:} To validate the effectiveness of our proposed core module (SMD module), we compared the performance of ST, SDIT and SMD on all datasets. SDPT does not have a modulation block, SDIT have a modulation encoder, and our STMDP features a modulation decoder (SMD). The results are shown in Table \ref{table2}. We achieved the best results in all the other three experiments. Especially in Can and Square tasks, we achieved 94\% and 80\% respectively. The results show that our SMD module has advantages in action trajectory generation. It can be clearly seen from Figure \ref{fig4}.


\begin{table*}[htbp]
\centering
\caption{The impact of the modulate block.}
\label{table2}
\begin{tabular}{c|c|c|c|c|c}
\hline
\multirow{2}{*}{Model} &\multirow{2}{*}{M-Block} & \multicolumn{4}{c}{Tasks}\\
\cline{3-6}

&& PushT & Can & Square & ToolHang \\
\hline

SDPT     &&	0.747 & 0.220	&0.040	&0.000\\ 
SDIT   &\checkmark&	0.700 & 0.920	&0.660	&\textbf{0.540}\\
\hline
\textbf{STMDP}  &\checkmark&\textbf{0.754} & \textbf{0.940}	& \textbf{0.800}	&\textbf{0.540}\\ 
\hline
\end{tabular}
\label{table_MAP}
\end{table*}


\section{Conclusion}

In this work, we proposed the Spiking Transformer Modulate Diffusion Policy Model, a new diffusion policy architecture based on the Spiking Transformer. This model effectively integrates SNNs with diffusion policy's capability to generate action trajectories. As a pioneering attempt based on the Spiking Transformer architecture in the diffusion policy, it achieves high-quality motion trajectory generation. The Spiking Modulate Decoder module we proposed is modulated based on the Transformer Decoder module, demonstrating the superiority of modulation based on the Decoder. Although the STMDP model has achieved good results, there is still room for further research on the selection of modulation modules and diffusion models. We will also conduct some experiments on the selection of DDIM parameters.

\subsection{Data Available}
The model of this research is one of the core and part of BrainCog Embot \cite{BrainCogEmbot} . BrainCog Embot is an Embodied AI platform under the Brain-inspired Cognitive Intelligence Engine (BrainCog) framework, which is an open-source Brain-inspired AI platform based on Spiking Neural Network.

\section{Acknowledgments}
This study was supported by the Postdoctoral Fellowship Program of CPSF (Grant No.GZC20232994), the Funding from Institute of Automation, Chinese Academy of Sciences(Grant No.E411230101), Special Research Assistant Program of the Chinese Academy of Sciences (Grant No.E4S9230501), Frontier Scientific Research Program of Chinese Academy of Sciences (ZDBS-LY-JSC013).

%
%
%
\newpage
\bibliographystyle{unsrt}
\bibliography{reference}

\end{document}